\pdfoutput=1

\documentclass[11pt]{article}
\usepackage[final]{ACL2023}

\usepackage{textalpha}

\usepackage{times}
\usepackage{latexsym}
\usepackage{array}

\usepackage{bbm}

\usepackage[T1]{fontenc}

\usepackage[utf8]{inputenc}
\usepackage{bm}

\usepackage{microtype}
\usepackage{pgfplots}
\pgfplotsset{compat=1.18}

\usepackage{graphicx}
\usepackage{inconsolata}
\usepackage{tabularx}
\usepackage{multirow}
\usepackage{amsmath} 
\usepackage{booktabs}

\title{LoRA-drop: Efficient LoRA Parameter Pruning based on Output Evaluation}

\author{Hongyun Zhou\thanks{\;\;Equal contribution} , Xiangyu Lu$^{*}$, Wang Xu, Conghui Zhu\thanks{\;\;Corresponding author}, Tiejun Zhao, Muyun Yang \\
  Faculty of Computing, Harbin Institute of Technology \\
  \{jameschou159,lu9995801,xwjim812\}@gmail.com, \{conghui,tjzhao,yangmuyun\}@hit.edu.cn}

\begin{document}
\maketitle
\begin{abstract}
Low-Rank Adaptation (LoRA) is currently the most commonly used Parameter-efficient fine-tuning (PEFT) method, it introduces auxiliary parameters for each layer to fine-tune the pre-trained model under limited computing resources.
However, it still faces resource consumption challenges during training when scaling up to larger models. 
Most previous studies have tackled this issue by using pruning techniques, which involve removing LoRA parameters deemed unimportant.
Nonetheless, these efforts only analyze LoRA parameter features to evaluate their importance, such as parameter count, size, and gradient.
In fact, the output of LoRA (product of LoRA parameter and hidden state), directly impacts the final results. 
Preliminary experiments indicate that a fraction of LoRA elements possesses significantly high output values, substantially influencing the layer output.
Motivated by the observation, we propose LoRA-drop.
Concretely, LoRA-drop evaluates the importance of LoRA based on the LoRA output.
Then we retain LoRA for important layers and the other layers share the same LoRA.
We conduct abundant experiments with models of different scales on NLU and NLG tasks. Results demonstrate that LoRA-drop can achieve performance comparable to full fine-tuning and LoRA, while retaining 50\% of the LoRA parameters on average. 

\end{abstract}

\section{Introduction}
\label{sec1}
\begin{figure}[t!]
    \centering
    \includegraphics{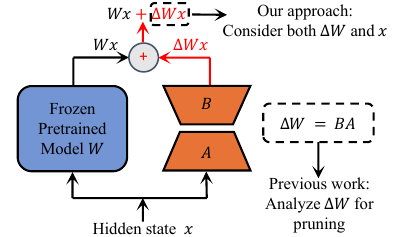}
    \caption{The diagram of LoRA. LoRA influences the pre-trained model through its output $\Delta \bm{Wx}$. This paper's method measures the importance of LoRA based on its output.}
    \label{fig.lora}
\end{figure}

Parameter-efficient fine-tuning methods have attracted more and more attention with the development of large language models (LLM)~\cite{li2021prefix,lester2021power}.
Among various PEFT methods, LoRA~\cite{hu2021lora} has been particularly prevalent in recent studies. 
LoRA freezes the pre-trained parameters and introduces auxiliary trainable parameters $\Delta \bm{W}$ for each layer as shown in Figure~\ref{fig.lora}.
LoRA significantly reduces the training cost while achieving impressive results. 

To further reduce the number of LoRA parameters being trained during efficient fine-tuning, previous studies employ pruning techniques that remove LoRA parameters deemed unimportant.
The core of these methods lies in how to evaluate the importance of parameters.
Sparse Adapter~\cite{he2022sparseadapter} evaluates the importance of LoRA based on different parameter features such as parameter count, parameter size, and parameter gradient. 
AdaLoRA~\cite{zhang2022adaptive} designs importance criteria based on the singular value decomposition (SVD) of $\Delta \bm{W}$ to prune unimportant singular values.
SoRA~\cite{ding2023sparse} prunes down-projection and up-projection matrices in LoRA by employing gate units and proximal gradient methods.
All of these efforts only focus on analyzing LoRA parameter $\Delta \bm{W}$ features to evaluate their importance, thereby reducing the parameters required for LoRA training.

\begin{figure*}[t!]
    \centering
    \includegraphics[width=\linewidth]{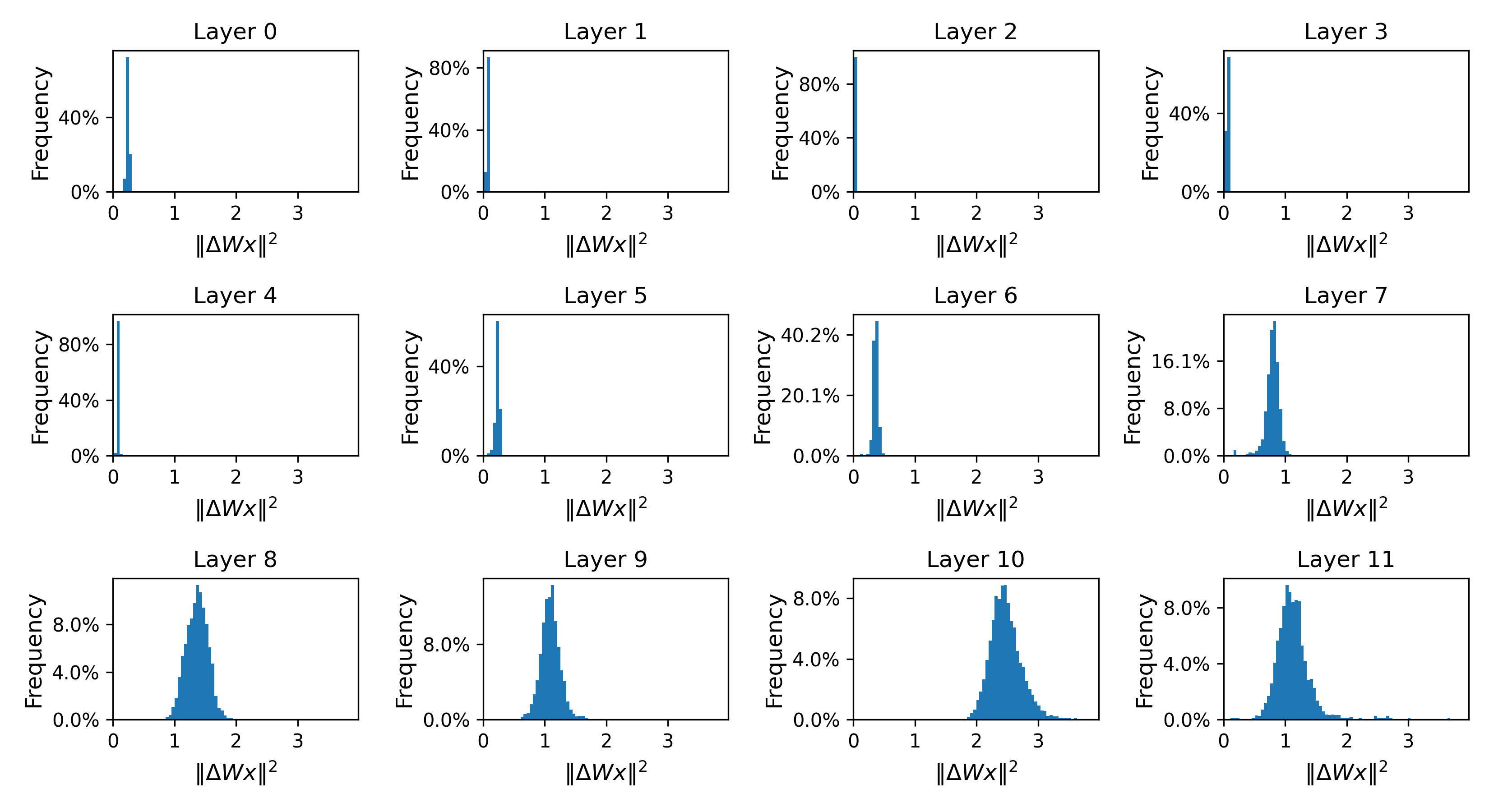}
    \caption{
    The frequency distribution of the squared norm of query LoRA output $\Delta \bm{W}_i\bm{x}_i$ on the RTE task. Each subplot represents the distribution of $\| \Delta \bm{W}_i\bm{x}_i \|^2$ for query LoRA from layers 0 to 11, where the x-axis denotes the magnitude of $\| \Delta \bm{W}_i\bm{x}_i \|^2$ for different inputs $\bm{x}_i$, and the y-axis represents the frequency of $\| \Delta \bm{W}_i\bm{x}_i \|^2$.}
    \label{fig.ydistribution}
\end{figure*}
In fact, the output of LoRA, which is related to the parameters and data, directly impacts the final results.
As shown in Figure~\ref{fig.lora}, the LoRA adds a bias term $\Delta \bm{Wx}$ in each layer of the pre-trained model.
Thus, the frozen model is fine-tuned by the bias term.
Intuitively, if the norm of $\Delta \bm{Wx}$ is large, the LoRA of this layer has an important impact on the frozen model.

We conducted an empirical study to analyze the distribution of LoRA output in LLMs. 
The findings derived from this study are presented in Section~\ref{sec2-pe}, revealing that the distribution of outputs from the LoRA of each layer is relatively concentrated. LoRA of some layers has little to no impact on specific tasks, while other layers exhibit more significant effects.
Thus, we could prune non-salient LoRA parameters.

Motivated by the observation, we propose LoRA-drop, which evaluates the importance of parameters by analyzing the LoRA output for each layer.
First, we sample specific task datasets and then utilize the sampled data to perform a limited number of updates to LoRA. 
The importance of LoRA for each layer is determined based on $\Delta \bm{Wx}$. 
Then, We retain the LoRA for layers with a large importance score, and the other layers share the same LoRA.
Finally, we fine-tune the model with fewer trainable parameters under the new LoRA setting, while minimizing performance degradation.

Our contributions are as follows:

\begin{itemize}
    \item We conducted empirical research, and the analysis indicates that the distribution of outputs from the LoRA of each layer is relatively concentrated. LoRA of some layers has little to no impact on specific tasks, while other layers exhibit more significant effects.
    \item We propose LoRA-drop, which evaluates the importance of LoRA for different layers and significantly reduces the parameter required during LoRA training while maintaining performance comparable to standard LoRA.
    \item We conduct comprehensive experiments on multiple NLU and NLG tasks with various sizes of pre-trained models. Numerous analysis experiments demonstrate the effectiveness of LoRA-drop.
\end{itemize}


\begin{figure*}
    \centering
    \includegraphics[width=\linewidth]{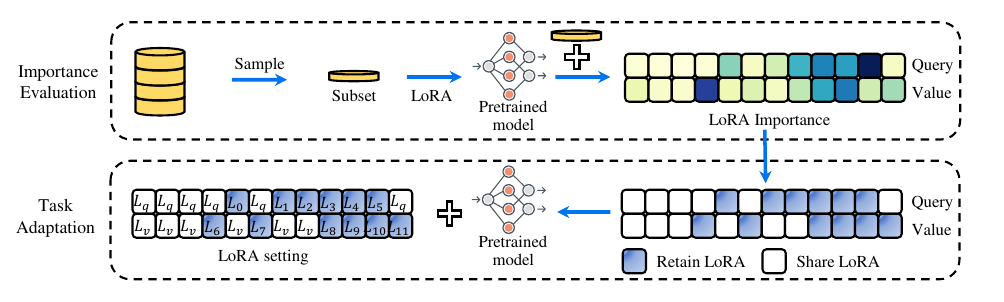}
    \caption{The overall workflow of LoRA-drop.}
    \label{fig:overall_workflow}
\end{figure*}

\section{Preliminary Experiment}\label{sec2-pe}
LoRA utilizes the product of two low-rank matrices to simulate incremental updates to a full-rank matrix. 
The pre-trained parameters are frozen during training and do not receive gradient updates, while the two low-rank matrices are trained. 
Let $\bm{W}_i$ denote the query/key/value matrix of $i$th Transformer layer and $\bm{x}_i$ denote the input of the $i$th Transformer.
The two low-rank matrices are $\bm{A}_i$ and $\bm{B}_i$.
Thus, the query/key/value vector is as follows:
\begin{equation}
\bm{h}_i=\bm{W}_{i}\bm{x}_i+\Delta \bm{W}_i\bm{x}_i=\bm{W}_{i}\bm{x}_i+\bm{B}_i\bm{A}_i\bm{x}_i 
\end{equation}
where $\Delta \bm{W}_i\bm{x}_i$ is the bias introduced by the LoRA modules.

Obviously, the $\Delta \bm{W}_i\bm{x}_i$ is the factor that directly influences the frozen pre-trained model. 
The larger $\Delta \bm{W}_i\bm{x}_i$, the greater the impact of LoRA on the pre-trained model, and consequently, the more important LoRA is. 
In fact, the $\Delta \bm{W}_i\bm{x}_i$ is related to the LoRA parameter and the hidden state, where the hidden state is computed from downstream task data through the preceding layers of the model.
However, previous work prunes LoRA by only analyzing its parameter features, ignoring the hidden state.

Preliminarily, we statistics the distribution of the LoRA output in each layer.
Specifically, we fine-tune the RoBERTa-base model with LoRA separately on the RTE and MRPC dataset, and analyze the distribution of the squared norm of the LoRA output $\Delta \bm{W}_i\bm{x}_i$ for each dataset.
We evaluate the impact of LoRA by computing the squared norm of $\Delta \bm{W}_i\bm{x}_i$.
Following the setting of \cite{hu2021lora}, the LoRA is added to the query and value matrix.
The distribution of query and value LoRA for RTE is shown in Figure~\ref{fig.ydistribution} and Figure~\ref{fig.rte_value}.
The distribution of query and value LoRA for MRPC is shown in Figure~\ref{fig:mrpc_query} and Figure~\ref{fig:mrpc_value}.

As observed, the squared norm distribution of $\Delta \bm{W}_i\bm{x}_i$ for each layer is highly concentrated, showing a peak Gaussian frequency distribution, which suggests stability.
Furthermore, Observations show that the squared norm of $\Delta \bm{W}_i\bm{x}_i$ for certain layers consistently remains close to zero, indicating that LoRA for these layers has almost no impact on the frozen model.
Conversely, some layers show a more significant impact on the frozen model.

Moreover, RTE and MRPC exhibit different distribution patterns.
It indicates that different layers play varying roles across different tasks.

This preliminary experiment demonstrates that we can prune the LoRA to reduce the number of trainable parameters.
LoRA with small $\Delta \bm{W}_i\bm{x}_i$ is insignificant, and can be pruned.

\section{Methodology}
\label{sec3}

In this section, we introduce LoRA-drop, a novel parameter-efficient fine-tuning method that prunes based on LoRA output. 
We have designed a process to quantify the importance of LoRA for different layers based on its output. 
Then, we retain the more important LoRA and replace the less important ones with a shared LoRA parameter, thereby reducing the number of parameters required for LoRA training while maintaining performance comparable to that of the standard LoRA.

Specifically, LoRA-drop consists of two parts: \textbf{Importance Evaluation} and \textbf{Task Adaptation}. The overall process of LoRA-drop is illustrated in Figure \ref{fig:overall_workflow}.

\subsection{Importance Evaluation}\label{section3-1}

This step evaluates the importance of LoRA for different layers, providing a reference for its retention strategy in the Task Adaptation step. 

Since the A and B matrices of LoRA are initialized with Kaiming and zero initialization, the initial output is all zeros. The output of LoRA becomes meaningful only after certain update steps.

So, we first perform stratified sampling on the downstream task dataset to obtain a subset $D_{s}$ of training data $D$. The sampling ratio is set to $\alpha$, where 0<$\alpha$<1. After that, the LoRA parameters are updated with several steps using this subset.

Next, we compute the sum of the squared norm of the LoRA output for each layer, denoted as $\bm{g}$, the $\bm{g}$ of the $i$-th layer LoRA as expressed in Equation~\ref{eq:gi}.

\begin{equation}
\bm{g}_i = \sum_{\bm{x} \in D_{s}} \|\Delta \bm{W}_i \bm{x}_i\|^2 \label{eq:gi}
\end{equation}

From section \ref{sec2-pe}, the magnitude of $g$ reflects the importance of LoRA. To better represent the relative importance of LoRA for each layer, we normalized $g$, resulting in the importance $I$ for each layer of LoRA.

\begin{equation}
\bm{I}_i = \frac{\bm{g}_i}{\sum_i \bm{g}_i}
\end{equation}

Thus, the importance of each layer of LoRA is bounded between 0 and 1, with a total sum of 1.

We find that sampling a small subset from the training data is able to obtain a LoRA importance distribution similar to that of the full dataset. This was verified by experiments in Section \ref{Analysis}. Our experiments' default value of $\alpha$ is set to 10\%.

\subsection{Task Adaptation}
This step sets the LoRA-drop fine-tuning strategy suitable for the downstream task based on the LoRA importance distribution. 

With the importance of LoRA for each layer, we sort the layers according to $\bm{I}_i$.
We select the layers from most to least important until the sum importance of the selected layer reaches a threshold $T$. 
In this paper, $T$ is set to 0.9 by default, and the value of $T$ is discussed in section \ref{Threshold_T}.

The LoRA of these selected layers will be retained during training, while a shared LoRA parameter will replace the LoRA of the other layers.
The hyper-parameter $T$ controls the number of the selected layers.
Finally, we fine-tune the model using the training dataset under the new LoRA setting.

\section{Experiments}
\label{sec4}

\begin{table*}[h!]
\centering
\scalebox{0.8}{
\begin{tabular}{l|c|ccccccccc}
\toprule
\begin{tabular}[c]{@{}l@{}}Model\\ RoBERTa-base\end{tabular} & \begin{tabular}[c]{@{}c@{}}\#Tr.\\  Params\end{tabular} & \begin{tabular}[c]{@{}c@{}}RTE\\ (Acc)\end{tabular} & \begin{tabular}[c]{@{}c@{}}MRPC\\ (Acc)\end{tabular} & \begin{tabular}[c]{@{}c@{}}STS-B\\ (Spea.)\end{tabular} & \begin{tabular}[c]{@{}c@{}}CoLA\\ (Matt.)\end{tabular} & \begin{tabular}[c]{@{}c@{}}SST-2\\ (Acc)\end{tabular} & \begin{tabular}[c]{@{}c@{}}QNLI\\ (Acc)\end{tabular} & \begin{tabular}[c]{@{}c@{}}MNLI\\ (Acc)\end{tabular} & \begin{tabular}[c]{@{}c@{}}QQP\\ (Acc)\end{tabular} & Avg.\\ 
\midrule
Full-FT* & 125M & $78.7$ & $\textbf{90.2}$ & $\textbf{91.2}$ & $\underline{63.6}$ & $\textbf{94.8}$ & $92.8$ & $\textbf{87.6}$ & $\textbf{91.9}$ & $\textbf{86.4}$ \\
LoRA & 0.29M & $\underline{80.8}_{\pm 1.5}$ & $89.1_{\pm 0.6}$ & $\textbf{91.2}_{\pm 0.2}$ & $62.4_{\pm 0.7}$ & $94.3_{\pm 0.3}$ & $\underline{93.0}_{\pm 0.2}$ & $\underline{87.5}_{\pm 0.2}$ & $\underline{90.3}_{\pm 0.1}$ & $86.1$ \\
SoRA & 0.21M & $79.7_{\pm 0.7}$ & $\underline{89.7}_{\pm 1.0}$ & $89.8_{\pm 0.1}$ & $\textbf{63.8}_{\pm 1.0}$ & $\textbf{94.8}_{\pm 0.4}$ & $92.4_{\pm 0.3}$ & $86.1_{\pm 0.1}$ & $88.9_{\pm 0.3}$ & $85.6$ \\
Sparse Adapter & 0.15M & $78.7_{\pm 1.1}$ & $88.0_{\pm 0.5}$ & $89.5_{\pm 0.4}$ &$60.1_{\pm 0.7}$ & $94.1_{\pm 0.1}$ & $92.8_{\pm 0.1}$ & $87.1_{\pm 0.2}$ & $89.6_{\pm 0.1}$ & $85.0$ \\
 VeRA & 0.03M & $78.0_{\pm 1.1}$ & $88.4_{\pm 0.1}$ & $89.8_{\pm 0.2}$ & $60.9_{\pm 0.5}$ & $93.7_{\pm 0.1}$ & $89.6_{\pm 0.1}$ & $83.7_{\pm 0.1}$ & $86.8_{\pm 0.1}$ & $83.9$ \\
 Tied-LoRA & 0.15M & $80.0_{\pm 0.9}$ & $89.1_{\pm 0.6}$ & $90.3_{\pm 0.1}$ & $62.0_{\pm 0.8}$ & $94.1_{\pm 0.3}$ & $91.6_{\pm 0.4}$ & $86.9_{\pm 0.1}$ & $89.7_{\pm 0.1}$ & $85.5$ \\
 LoRA-drop~(ours) & 0.15M & $\textbf{81.4}_{\pm 0.5}$ & $89.5_{\pm 0.5}$ & $\underline{91.0}_{\pm 0.1}$ & $62.9_{\pm 0.2}$ & $\underline{94.5}_{\pm 0.2}$ & $\textbf{93.1}_{\pm 0.1}$ & $87.3_{\pm 0.2}$ & $90.1_{\pm 0.1}$ & $\underline{86.2}$ \\
\bottomrule
\end{tabular}}
\caption{Results of the RoBERTa-base with different training strategies on the GLUE benchmark. The results are averaged from three seeds to produce solid results. The subscript is the standard deviation. Bold and underlined indicate the first and second best results in the corresponding regime. \#Tr. refers to trainable. * refers to the results directly from their original paper, in which Full-FT is derived from \cite{liu2019roberta}.}
\label{tab:base}
\end{table*}

\begin{table*}[t!]
\centering
\scalebox{0.8}{
\begin{tabular}{l|c|ccccc}
\toprule
Model                          & \#Tr.                       & \multicolumn{1}{c}{E2E}    & \multicolumn{1}{c}{DART}   & \multicolumn{1}{c}{Dialogsum} & \multicolumn{1}{c}{GSM8K} & \multirow{2}{*}{Avg.} \\
\multicolumn{1}{l|}{Llama2 7b} & \multicolumn{1}{l|}{Params} & \multicolumn{1}{c}{(BLEU)} & \multicolumn{1}{c}{(BLEU)} & \multicolumn{1}{c}{(ROUGE)} & \multicolumn{1}{c}{(Acc)} \\ 
\midrule
Full-FT          & 6.6B   & $55.65$           & $\textbf{59.68}$ & $40.77$           & $31.16$           & $46.82$ \\
LoRA             & 0.13B  & $56.38$           & $58.51$          & $\textbf{41.03}$  & $34.04$           & $47.49$ \\
LoRA-drop (ours) & 0.09B  & $\textbf{57.06}$  & $58.82$          & $40.68$           & $\textbf{34.50}$  & $\textbf{47.77}$ \\ 
\bottomrule
\end{tabular}}
\caption{Results of Llama2-7b with different training strategies on two table2text datasets including E2E and DART, the summarization dataset Dialogsum, and the mathematical reasoning dataset GSM8K. For all the scores, BLEU, ROUGE, and Acc, higher is better.}
\label{tab:NLG}
\end{table*}

\subsection{Setup}
\paragraph{Datasets.}
We evaluate our model on both Natural Language Understanding (NLU) and Natural Language Generation (NLG) tasks. 

For NLU, we evaluate our method on the GLUE benchmark~\cite{wang2018glue}, which consists of eight datasets: CoLA, SST-2, MRPC, QQP, STS-B, MNLI, QNLI, and RTE. 
We use Matthew's correlation coefficient, Spearman's correlation coefficient, and overall accuracy (for both matched and mismatched sentences) to evaluate the CoLA, STS-B, and MNLI datasets. 
For the remaining datasets, we apply accuracy as the evaluation metric. 

The NLG tasks in our experiments include the table-to-text datasets E2E~\cite{dusek.etal2020:csl} and DART~\cite{nan-etal-2021-dart}, the summarization dataset DialogSum~\cite{chen-etal-2021-dialogsum}, as well as the Mathematical Reasoning dataset GSM8K~\cite{cobbe2021gsm8k}. 
We use BLEU~\cite{papineni2002bleu}, ROUGE~\cite{lin-2004-rouge}, and accuracy to evaluate the E2E(\&DART),  DialogSum, and GSM8K datasets.

\paragraph{Baselines.}
The following methods are chosen as baselines:
\textbf{FULL-FT} updates all model parameters which need a lot of computing resources. 
\textbf{LoRA}~\cite{hu2021lora} represents the original LoRA method.
\textbf{Sparse Adapter}~\cite{he2022sparseadapter} applies typical pruning methods to LoRA and reduces the trainable parameters.
\textbf{VeRA}~\cite{kopiczko2024vera} shares and freezes randomly initialized LoRA and introduces trainable vectors for each layer to reduce the parameters of LoRA.
\textbf{Tied-LoRA}~\cite{renduchintala2023tied} makes the frozen LoRA in VeRA trainable.
\textbf{SoRA}~\cite{ding2023sparse}uses a gate unit with proximal gradient methods to control LoRA's sparsity.

\paragraph{Models \& Implementation.}
To evaluate the effectiveness of our method on various models, we conduct experiments on RoBERTa-base, RoBERTa-large\cite{liu2019roberta}, and Llama2-7b\cite{touvron2023llama}. 
We conduct NLU experiments on the GLUE benchmark using all three models. 
We performed 3 runs with different random seeds for each dataset and recorded the best results for each run.
The average results and the standard deviation are calculated.

To evaluate the effectiveness of our method on generation tasks, we conduct NLG experiments using the Llama2-7b on the table2text datasets: E2E and DART, the summarization dataset DialogSum, as well as the Mathematical Reasoning dataset GSM8K. 

The hyperparameter settings for each baseline and LoRA-drop can be found in Section~\ref{sec_detail}.

\subsection{Main Results}\label{sec42}
The main results of RoBERTa-base with different training methods on the GLUE benchmark are shown in Table~\ref{tab:base}.
It is noted that our motivation is to reduce the number of trainable parameters while ensuring that the performance does not degrade, or even improve.
As shown in Table~\ref{tab:base}, with an approximately 50\% reduction in standard LoRA parameters, our proposed LoRA-drop achieves an average score of 86.2, on par with Full-FT (86.4) and LoRA (86.1). 
This indicates the effectiveness of our proposed LoRA-drop, which outperforms LoRA by 0.1 scores while reducing training parameters.

Moreover, LoRA-drop achieves 0.6, 1.2, 2.3, and 0.7 improvements in average scores compared to the four baselines: SoRA, Sparse Adapter, VeRA, and Tied-LoRA respectively. Although all four methods effectively reduce LoRA parameters, their performance drops significantly. 
The results demonstrate that LoRA-drop is a superior strategy for evaluating the importance of trainable parameters and reducing less important ones, thereby enhancing parameter efficiency.

The results of RoBERTa-large and Llama2-7b with different training strategies on the GLUE benchmark are presented in Table~\ref{tab:large} and Table~\ref{tab:llama}.
It is noted that we use Llama2-7b to obtain the token representation rather than generate the answer.
On both models, our method utilizes about 52\% of the standard LoRA parameters and achieves average scores of 89.1 and 89.3 for RoBERTa-large and Llama2-7b respectively, outperforming LoRA and Full-FT. This demonstrates the effectiveness of our method across models of different scales.

The results of NLG tasks, including table2text, summarization, and mathematical reasoning, are shown in Table~\ref{tab:NLG}. On Llama2-7b, our method achieves results on par with the Full-FT and LoRA while using approximately 68\% of the original LoRA parameters for all three tasks. Additionally, the average score of our method (47.77) exceeds that of Full-FT (46.82) and LoRA (47.49). This confirms the effectiveness of our method across both NLU and NLG. 

\begin{table*}[t!]
\centering
\scalebox{0.8}{
\begin{tabular}{l|ccccccccc}
\toprule
\begin{tabular}[c]{@{}l@{}}Model\\ (RoBERTa-base)\end{tabular} & \begin{tabular}[c]{@{}c@{}}RTE\\ (Acc)\end{tabular} & \begin{tabular}[c]{@{}c@{}}MRPC\\ (Acc)\end{tabular} & \begin{tabular}[c]{@{}c@{}}STS-B\\ (Spea.)\end{tabular} & \begin{tabular}[c]{@{}c@{}}CoLA\\ (Matt.)\end{tabular} & \begin{tabular}[c]{@{}c@{}}SST-2\\ (Acc)\end{tabular} & \begin{tabular}[c]{@{}c@{}}QNLI\\ (Acc)\end{tabular} & \begin{tabular}[c]{@{}c@{}}MNLI\\ (Acc)\end{tabular} & \begin{tabular}[c]{@{}c@{}}QQP\\ (Acc)\end{tabular} & Avg. \\
\midrule
 LoRA                 & $79.4$ & $89.2$ & $91.0$ & $63.1$ & $94.6$ & $92.7$ & $87.6$ & $90.3$ & $86.0$ \\
 LoRA(large $I$) & $72.2$ & $77.5$ & $85.9$ & $58.9$ & $92.9$ & $73.6$ & $71.2$ & $82.6$ & $76.9$ \\
 LoRA(small $I$) & $47.7$ & $69.9$ & $49.6$ & $23.5$ & $88.2$ & $55.4$ & $32.2$ & $63.9$ & $53.8$ \\
\bottomrule
\end{tabular}}
\caption{Verification of Importance Evaluation Method. The data in the table represents the results from a single run with the same random seed.
LoRA (large $I$) retains the few LoRAs with the highest $I$ values, while LoRA (small $I$) retains the few with the lowest $I$ values. The number retained is consistent with the LoRA-drop setting in Table~\ref{tab:base}.}
\label{tab:half-inference}
\end{table*}

\subsection{Analysis}\label{Analysis}

\begin{figure*}[h!]
    \centering
    \includegraphics[width=\linewidth]{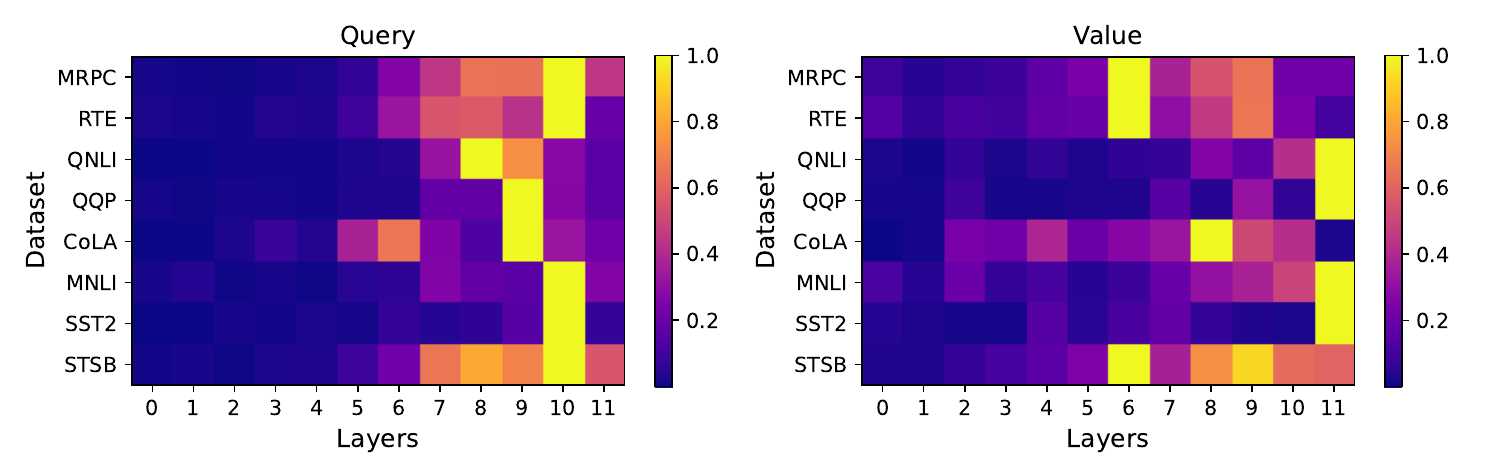}
    \caption{LoRA Importance Distribution in Different Downstream Task Data. To unify the importance scales across different datasets, we divide the importance of each dataset by its maximum value so that the importance of the most important layer of LoRA in that dataset is 1.}
    \label{fig:different datasets}
\end{figure*}

\paragraph{The value of LoRA output indicated the importance.}
As described in Section~\ref{section3-1}, the importance evaluation step quantifies the importance of LoRA based on its output.
In this section, we verify the effectiveness of the output-based evaluation method.
Specifically, we first perform standard LoRA fine-tuning and obtain the importance score.
Based on this score, we retain either the largest or the smallest of the LoRA layers for inference, the number of retained LoRA is consistent with the number retained by LoRA-drop in Section~\ref{sec42}.
We then evaluate these two settings, and the final results are presented in Table~\ref{tab:half-inference}.

It is evident that when only approximately half of the LoRA modules are retained, the model's performance decreases significantly. When we retain the LoRA modules with larger $I$, the performance is substantially better than those with smaller $I$. 
This indicates that the LoRA-drop method's layer-specific LoRA Importance Evaluation is effective. 
LoRA with a larger squared norm output indeed has a greater contribution to the model's fine-tuning performance.

\paragraph{Distribution of LoRA importance varies across different tasks.}
The insight of our approach is to derive LoRA importance adapted to the distribution of different downstream task data, enabling the simplification of LoRA parameters. 
To further validate the rationality of this insight, we plot heatmaps illustrating the distribution of LoRA importance $\bm{I}$ for eight different datasets in GLUE on RoBERTa-base and Llama2.

The results are presented in Figure~\ref{fig:different datasets} and Figure~\ref{fig:llama different datasets}.
We observe that the importance distributions differ across datasets, indicating that the importance assigned by LoRA is data-dependent. 


\paragraph{The influence of LoRA share.} 
In our method, the layers with low importance are shared with the same LoRA parameters. 
We investigate the influence after the LoRA parameters are shared.
Following the LoRA share operation on the RoBERTa-base model trained on 20\% of the RTE training set data for 4 epochs, we plot the importance distribution for each layer of the model.

The results of query and value distribution are shown in Figure~\ref{fig.rte_query_share} and Figure~\ref{fig.rte_value_share}.
It shows that the importance distribution of LoRA for each layer remains almost consistent with the original LoRA after the LoRA parameters are shared. 
This suggests that the sharing LoRA for layers with low importance does not affect the importance distribution of other layers, thereby maintaining good performance.

\begin{figure}[htb!]
    \centering
    \includegraphics[width=\linewidth]{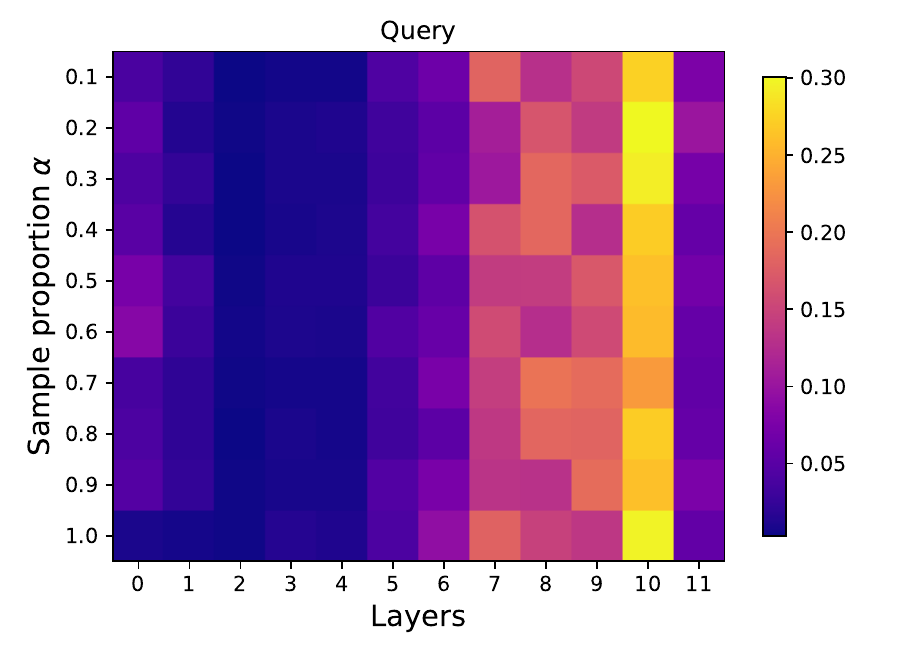}
    \caption{Importance distribution of LoRA for query in RTE under different sample proportions. Each point on the heatmap represents the importance $I_{i}$ of the query LoRA in layer $i$ under $\alpha$ sample proportion.}
    \label{fig:sample proportion query}
\end{figure}

\begin{table*}[h]
\centering
\scalebox{0.8}{
\begin{tabular}{c|cc|ccccc}
\toprule
\multirow{2}{*}{\centering Threshold} & \multicolumn{2}{c|}{Avg. layer num}
 & \multirow{2}{*}{\begin{tabular}[c]{@{}c@{}}RTE\\ (ACC)\end{tabular}} & \multirow{2}{*}{\begin{tabular}[c]{@{}c@{}}CoLA\\ (Matt.)\end{tabular}} & \multirow{2}{*}{\begin{tabular}[c]{@{}c@{}}QNLI\\ (ACC)\end{tabular}} & \multirow{2}{*}{\begin{tabular}[c]{@{}c@{}}QQP\\ (ACC)\end{tabular}} & \multirow{2}{*}{Avg.} \\
 & W\_query & W\_value & & & & & \\
\midrule
1(LoRA)    & 12           & 12            & 82.3                  & 61.9                  & 93.1                 & 90.4                 & 82.0                        \\
0.95       & 6            & 9             & 83.0                  & 62.6                  & 93.1                 & 90.2                 & 82.2                        \\
0.9        & 5            & 7             & 81.9                  & 63.1                  & 93.2                 & 90.2                 & 82.1                        \\
0.8        & 5            & 5             & 80.9                  & 63.1                  & 93.2                 & 89.6                 & 81.7                        \\
0.7        & 4            & 4             & 78.3                  & 62.1                  & 92.5                 & 89.3                 & 80.6                        \\ 
\bottomrule
\end{tabular}}
\caption{The influence of the threshold $T$ and its equivalent average number of layers.}
\label{tab:T}
\end{table*}

\begin{table*}[]
\centering
\scalebox{0.8}{
\begin{tabular}{l|ccccccccc}
\toprule
\begin{tabular}[c]{@{}l@{}}Model\\ (RoBERTa-base)\end{tabular} & \begin{tabular}[c]{@{}c@{}}RTE\\ (Acc)\end{tabular} & \begin{tabular}[c]{@{}c@{}}MRPC\\ (Acc)\end{tabular} & \begin{tabular}[c]{@{}c@{}}STS-B\\ (Spea.)\end{tabular} & \begin{tabular}[c]{@{}c@{}}CoLA\\ (Matt.)\end{tabular} & \begin{tabular}[c]{@{}c@{}}SST-2\\ (Acc)\end{tabular} & \begin{tabular}[c]{@{}c@{}}QNLI\\ (Acc)\end{tabular} & \begin{tabular}[c]{@{}c@{}}MNLI\\ (Acc)\end{tabular} & \begin{tabular}[c]{@{}c@{}}QQP\\ (Acc)\end{tabular} & Avg. \\
\midrule
 LoRA-drop* & $81.9$ & $90.0$ & $91.1$ & $63.1$ & $94.7$ & $93.2$ & $87.5$ & $90.2$ & $86.5$ \\
 LoRA-drop(w/o share) & $80.4$ & $88.9$ & $90.7$ & $62.8$ & $94.1$ & $92.9$ & $86.9$ & $89.7$ & $85.8$ \\
 LoRA-drop($\Delta Wx$ inv) & $79.1$ & $89.7$ & $90.4$ & $60.5$ & $94.3$ & $92.9$ & $87.3$ & $89.9$ & $85.5$ \\
 LoRA-drop(random) & $79.1$ & $89.2$ & $90.2$ & $62.0$ & $94.6$ & $92.7$ & $86.9$ & $89.8$ & $85.6$ \\
 LoRA-drop(top $k$) & $81.9$ & $89.2$ & $90.7$ & $62.3$ & $94.5$ & $93.0$ & $86.8$ & $89.8$ & $86.0$ \\
\bottomrule
\end{tabular}}
\caption{Ablation experiments.}
\label{tab:ablation}
\end{table*}

\paragraph{The influence of sample proportion.} 
We investigate the influence of the sample proportion when calculating the importance of LoRA. 
We sample ten different-sized datasets from the RTE dataset with sampling ratios from 10\% to 100\%.
We train the RoBERTa-base model using LoRA for the same number of steps and obtain the LoRA importance for each sample proportion.

The results of LoRA for Query and Value are shown in Figure~\ref{fig:sample proportion query} and Figure~\ref{fig.sample}.
As the training data increases, the importance order of each layer remains consistent. For LoRA applied to the query matrices, the 10th layer has always been the most important, while the importance of layers 7, 8, and 9 maintains a consistently high level of importance.
Indicating that this operation is insensitive to the size of the sampled data and exhibits robustness.

\paragraph{Selection of threshold $T$.}\label{Threshold_T}
LoRA-drop introduces a hyper-parameter $T$ to control the number of selected layers.
We select four datasets from GLUE to analyze the impact of threshold $T$.

The results are shown in Table~\ref{tab:T}.
When T is set to 1, all layers are preserved, representing the standard LoRA method.
When $T$ is less than 0.9, the model performance increases with $T$,  at this time, LoRA modules with higher importance are selected. When $T$ equals 0.9, approximately half of the layers' LoRA are selected on average. 
If $T$ continues to increase, the newly added LoRA modules have lower importance, and the model performance no longer significantly improves. Hence in our experiments, we default to setting $T$ as 0.9.

\subsection{Ablation Study}
In this subsection, we conduct ablation experiments to verify the following two questions:
\begin{itemize}
    \item Q1: Is replacing LoRA for layers with small $\bm{I}$ with shared parameters better than directly removing them in the task adaptation step?
    \item Q2: Is retaining LoRA with large $\bm{I}$ in the task adaptation step reasonable? 
\end{itemize}

To answer these two questions, we compare LoRA-drop with the following variants on the RoBERTa-base model, where $k$ refers to the number of LoRA retained by LoRA-drop.

\textbf{LoRA-drop (w/o share)} directly removes the low-importance layers of LoRA without using additional shared parameters in the Task Adaptation step.
As opposed to LoRA-drop, \textbf{LoRA-drop ($\Delta \bm{Wx}$ inv)} replace high-importance layers of LoRA with shared LoRA and retain the other LoRA. 
\textbf{LoRA-drop (random)} randomly selects $k$ layers that retain LoRA parameters.
\citet{houlsby2019parameter} found that lower layers often have a small impact on performances, so \textbf{LoRA-drop (top $k$)} selects the top $k$ layers of the 12-layer model.
We experiment with these four settings on the validation set of the GLUE benchmark.

The results are shown in Table \ref{tab:ablation}.

\textbf{Regarding Q1}, directly removing less important LoRA parameters, i.e., the LoRA-drop (w/o share) setting, performs worse across all tasks than LoRA-drop, with an average reduction of 0.7 scores. 

This indicates that sharing a LoRA among the layers with low importance is necessary to achieve better fine-tuning results compared to directly removing them. 

\textbf{Regarding Q2}, the $\Delta \bm{Wx}$ inv setting achieved the worst average performance, slightly worse than the random setting. This indicates that LoRA with smaller $\bm{I}$ contributes less to model performance improvement. The top $k$ setting, which empirically retains the top $k$ layers, performed well but had an average performance gap of 0.5 scores compared to the LoRA-drop.

LoRA-drop yields better performance compared to all the other three variants. It confirms the reasonableness of retaining the LoRA of layers with significant importance and further validates the effectiveness of the method proposed in this paper for evaluating the importance of LoRA.

\section{Related Work}

Parameter Efficient Fine-Tuning~(PEFT) is the mainstream method for the current fine-tuning of pre-trained models, which can be broadly categorized into additive methods, selective methods, and reparameterized~\cite{han2024parameter}.

\subsection{Additive Methods}  
Additive methods inject new trainable modules or parameters into pre-trained models.
During fine-tuning for a specific downstream task, only the weights of these newly added modules are updated. 

Adapter~\cite{houlsby2019parameter} involves inserting small adapter layers within Transformer blocks. 
There are two ways to inject adapters into pre-trained models: Serial Adapter~\cite{houlsby2019parameter} adds two adapter modules in each Transformer block. 
Parallel Adapter~\cite{he2021towards} transforms the serial adapter layers into parallel side networks.
Adapter Drop~\cite{ruckle2021adapterdrop} empirically removes lower-layer Adapters considered to have a small impact on task performance. 

Soft Prompt uses continuous embedding of soft prompts instead of optimizing discrete token representations through in-context learning.
Prefix-tuning~\cite{li2021prefix} inserts trainable vectors prepended to keys and values at all Transformer layers. 
P-tuning~\cite{liu2021gpt} and Prompt-tuning~\cite{lester2021power} only insert trainable vectors at the initial word embedding layer.

\subsection{Selective Methods}  
Selective methods make a small subset of parameters in the pre-trained model trainable while keeping the rest frozen. 
Diff pruning~\cite{guo2021parameter} employs a learnable binary mask on model weights.
BitFit \cite{zaken2022bitfit} only fine-tunes the bias parameters of each FFN, achieving competitive results for small models.
\citet{lee2019would} fine-tunes only the last quarter of BERT and RoBERTa's final layer, achieving 90\% of the performance of full fine-tuning. 
HiFi \cite{gui-xiao-2023-hifi} fine-tunes attention heads that are highly informative and strongly correlated for a specific task.

\subsection{Reparameterized Methods} 
In the context of PEFT, reparameterization often involves constructing a low-rank parameterization to enhance parameter efficiency during training.

LoRA~\cite{hu2021lora} introduces low-rank matrices during fine-tuning and can merge with pre-trained weights before inference.
There are many derivative works based on LoRA. 
QLoRA~\cite{dettmers2023qlora} quantifies the parameters of large models doubly, significantly reducing memory usage.
AdaLoRA~\cite{zhang2022adaptive} transforms the low-rank matrices in LoRA into SVD matrices $P\Lambda Q$. During training, the singular values are iteratively pruned. 
SoRA~\cite{ding2023sparse} eliminates the matrix orthogonality premise of $P$ and $Q$ in AdaLoRA and instead applies a gating unit between them.
Sparse Adapter~\cite{he2022sparseadapter} enhances the parameter efficiency of LoRA and other Adapters using network pruning methods. 
S2-LoRA \cite{liu2023sparsely} shares the LoRA parameters, and introduces trainable scaling vectors with inter-layer variations. 
VeRA \cite{kopiczko2024vera} and Tied-LoRA \cite{renduchintala2023tied}, further reduce the parameter count by sharing parameters for all layers and modules of LoRA. 
DoRA~\cite{liu2024dora} uses LoRA for directional updates, enhancing learning capacity and training stability.

\section{Conclusion}
In this paper, we propose a new parameter-efficient fine-tuning method LoRA-drop based on LoRA.
our motivation is to reduce the number of trainable parameters during fine-tuning while ensuring that the performance does not degrade, or even improve.
Concretely, we calculate the importance of LoRA for each layer based on the output. 
The LoRA parameters of layers with large importance are retained and the other layers share the same parameter, resulting in a significant reduction in the number of parameters that need to be trained compared to the original LoRA. 
Abundant experiments on multiple NLU and NLG datasets show that LoRA-drop can achieve comparable results with origin LoRA with 50\% of LoRA parameters.

\section*{Limitations}
Currently, our method operates on the LoRA structure as a whole, with a relatively coarse granularity. Future work will refine this method to a finer granularity. While this technique reduces the number of training parameters during LoRA training, it does not decrease the inference cost.
Pruning increases the model's complexity, making it more difficult to identify the sources of issues when performance falls short of expectations. This, in turn, complicates the processes of debugging and error analysis.

\bibliographystyle{acl_natbib}
\bibliography{custom}

\clearpage
\appendix
\section{Appendix}
\label{sec:appendix}

\subsection{Implementation Details}
\label{sec_detail}
Our LoRA configuration aligns with the experimental setup of \cite{hu2021lora}, where LoRA is applied to the query and value matrices in each self-attention module. We each use a shared LoRA in place of the low-importance query LoRA and value LoRA.

The low-rank matrix \textbf{$A$} of the LoRA architecture is initialized using Kaiming initialization \cite{he2015delving}, while matrix \textbf{$B$} is initialized with zeros. Unless specified otherwise, the default rank for LoRA is set to 8.

We conducted NLU experiments on the GLUE benchmark using RoBERTa-base \cite{liu2019roberta}. 
The data sampling ratio $\alpha$ is set to 0.1, the number of training epochs $n$ is set to 3, and the threshold $T$ for LoRA-drop is set to 0.9. 
To ensure consistency in the trainable parameter count between the baseline and our method, we set the sparsity rate of the Sparse Adapter to 0.5. We set the pruning method of the Sparse Adapter to the performance-optimal SNIP~\cite{2018SNIP}.
The rank of Tied-LoRA is set to 56. 
The design characteristics of the VeRA method determine that its trainable parameter count cannot reach the same order of magnitude as LoRA; 
otherwise, VeRA would no longer be a low-rank matrix. Therefore, we set the rank of VeRA to 512 based on the best hyperparameters provided in the original paper.

To evaluate the effectiveness of our method on generation tasks, we conducted NLG experiments using the Llama2 7b on the table2text datasets: E2E and DART, the summarization dataset DialogSum, as well as the mathematical reasoning dataset GSM8K. For all three tasks, we set the rank of LoRA to 64. 
It is worth noting that, in the NLG experiment we applied LoRA to the query, key, value, and output matrices in Attention, and up and down matrices in MLP, as we found that only fine-tuning the query and value matrices with LoRA would cause significant performance degradation.

\begin{figure*}
    \centering
    \includegraphics[width=\linewidth]{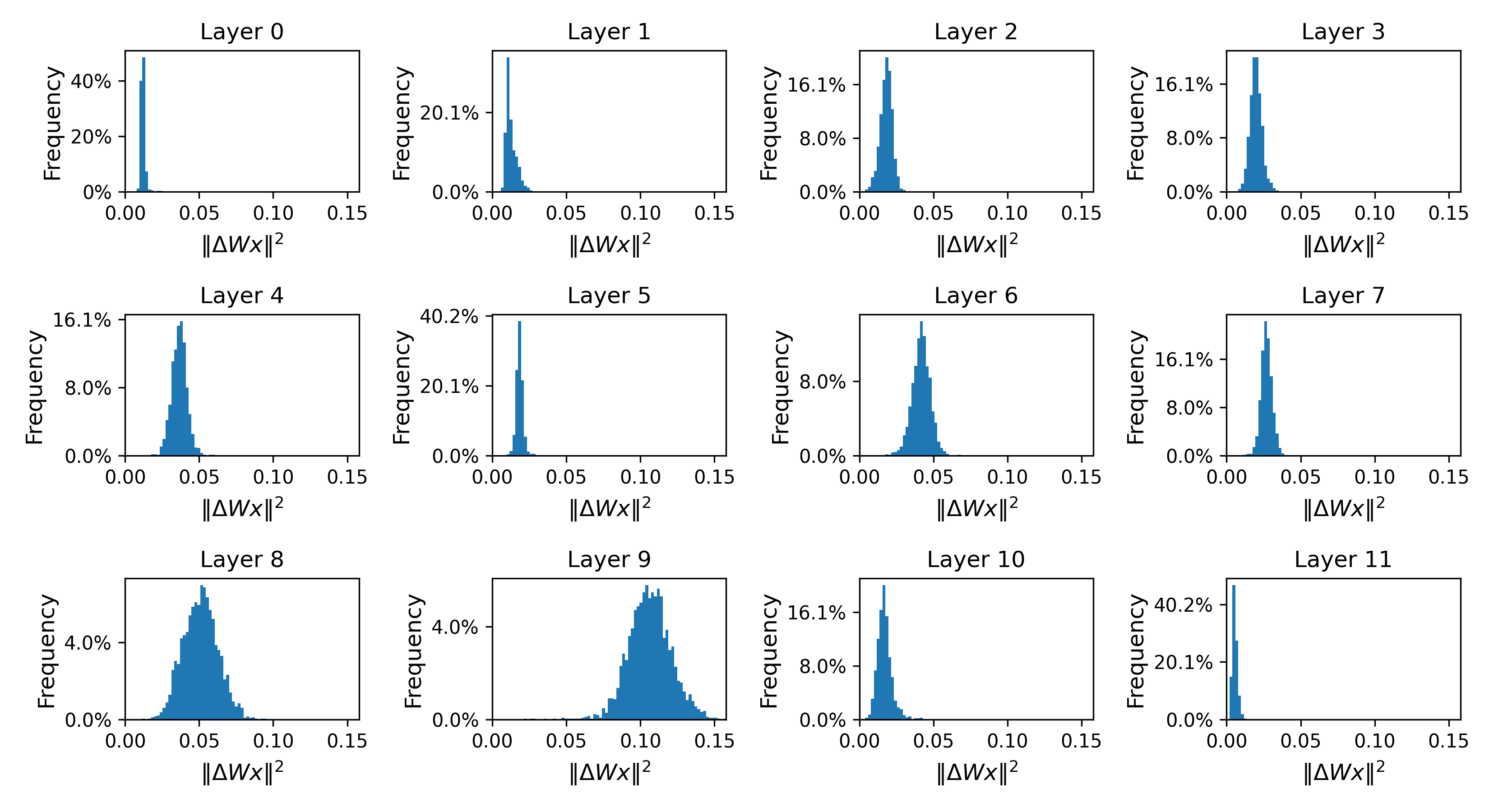}
    \caption{The frequency distribution of the squared norm of value LoRA output $\Delta \bm{W}_i\bm{x}_i$ after fine-tuning on the RTE task.}

    \label{fig.rte_value}
\end{figure*}
\begin{figure*}
    \centering
    \includegraphics[width=\linewidth]{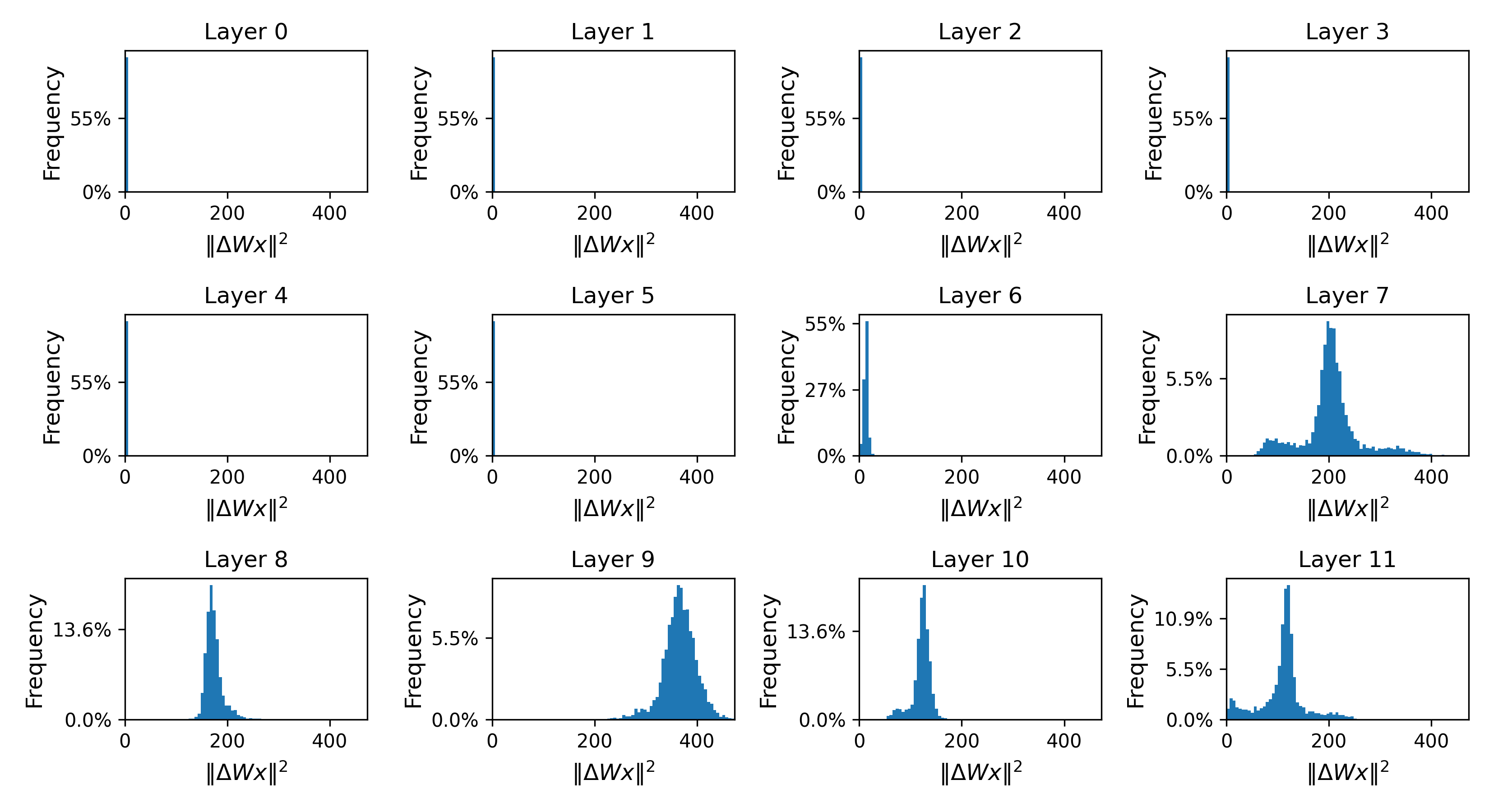}
    \caption{The frequency distribution of the squared norm of query LoRA output $\Delta \bm{W}_i\bm{x}_i$ after fine-tuning on the MRPC task.}
    \label{fig:mrpc_query}
\end{figure*}

\begin{figure*}
    \centering
    \includegraphics[width=\linewidth]{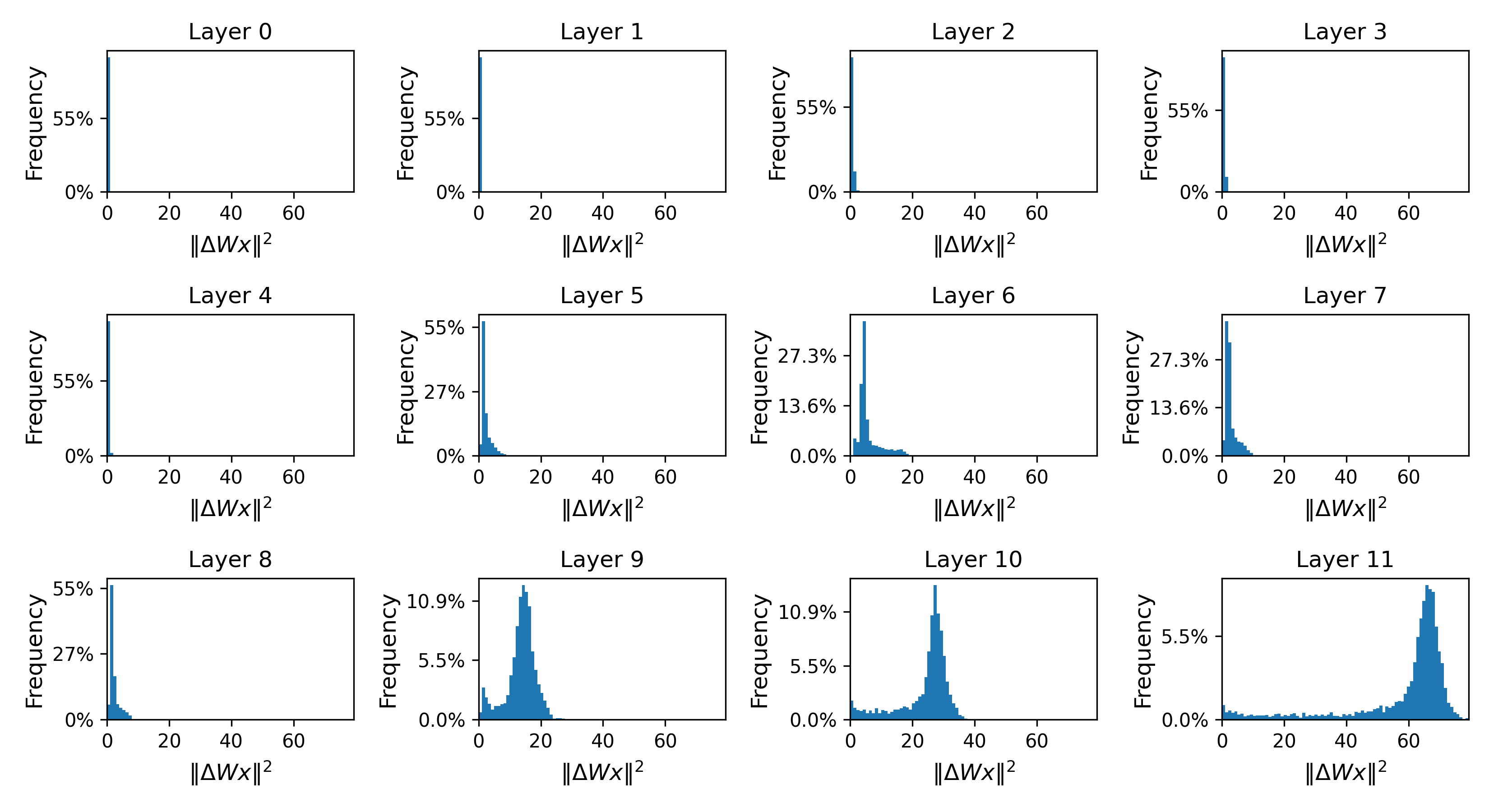}
    \caption{The frequency distribution of the squared norm of value LoRA output $\Delta \bm{W}_i\bm{x}_i$ after fine-tuning on the MRPC task.}
    \label{fig:mrpc_value}
\end{figure*}

\begin{table*}[]
\centering
\scalebox{0.8}{
\begin{tabular}{l|c|ccccccccc}
\toprule
\begin{tabular}[c]{@{}l@{}}Model\\ RoB-large\end{tabular} & \begin{tabular}[c]{@{}c@{}}\#Tr.\\  Params\end{tabular} & \begin{tabular}[c]{@{}c@{}}RTE\\ (Acc)\end{tabular} & \begin{tabular}[c]{@{}c@{}}MRPC\\ (Acc)\end{tabular} & \begin{tabular}[c]{@{}c@{}}STS-B\\ (Spea.)\end{tabular} & \begin{tabular}[c]{@{}c@{}}CoLA\\ (Matt.)\end{tabular} & \begin{tabular}[c]{@{}c@{}}SST-2\\ (Acc)\end{tabular} & \begin{tabular}[c]{@{}c@{}}QNLI\\ (Acc)\end{tabular} & \begin{tabular}[c]{@{}c@{}}MNLI\\ (Acc)\end{tabular} & \begin{tabular}[c]{@{}c@{}}QQP\\ (Acc)\end{tabular} & Avg. \\
\midrule
 Full-FT* & 355M & $86.6$ & $\textbf{90.9}$ & $\textbf{92.4}$ & $\underline{68.0}$ & $\textbf{96.4}$ & $94.7$ & $90.2$ & $\textbf{92.2}$ & $\underline{88.9}$ \\
 LoRA & 0.79M & $\underline{88.5}_{\pm 0.7}$ & $\underline{90.1}_{\pm 0.8}$ & $\textbf{92.4}_{\pm 0.1}$ & $67.8_{\pm 1.3}$ & $96.0_{\pm 0.1}$ & $\underline{94.8}_{\pm 0.1}$ & $\underline{90.6}_{\pm 0.0}$ & $\underline{91.4}_{\pm 0.1}$ & $\underline{88.9}$ \\
 LoRA-drop (ours)& 0.41M & $\textbf{88.8}_{\pm 0.7}$ & $89.9_{\pm 0.3}$ & $\underline{92.2}_{\pm 0.1}$ & $\textbf{68.5}_{\pm 1.7}$ & $\underline{96.2}_{\pm 0.1}$ & $\textbf{94.9}_{\pm 0.1}$ & $\textbf{90.7}_{\pm 0.1}$ & $91.3_{\pm 0.5}$ & $\textbf{89.1}$ \\
\bottomrule
\end{tabular}}
\caption{The performance of the RoBERTa-large on GLUE benchmark. * refers to the results directly from their original paper, in which Full-FT is derived from \cite{liu2019roberta}.}
\label{tab:large}
\end{table*}

\begin{table*}[]
\centering
\scalebox{0.8}{
\begin{tabular}{l|c|ccccccccc}
\toprule
\begin{tabular}[c]{@{}l@{}}Model\\ Llama2 7b\end{tabular} & \begin{tabular}[c]{@{}c@{}}\#Tr.\\  Params\end{tabular} & \begin{tabular}[c]{@{}c@{}}RTE\\ (Acc)\end{tabular} & \begin{tabular}[c]{@{}c@{}}MRPC\\ (Acc)\end{tabular} & \begin{tabular}[c]{@{}c@{}}STS-B\\ (Spea.)\end{tabular} & \begin{tabular}[c]{@{}c@{}}CoLA\\ (Matt.)\end{tabular} & \begin{tabular}[c]{@{}c@{}}SST-2\\ (Acc)\end{tabular} & \begin{tabular}[c]{@{}c@{}}QNLI\\ (Acc)\end{tabular} & \begin{tabular}[c]{@{}c@{}}MNLI\\ (Acc)\end{tabular} & \begin{tabular}[c]{@{}c@{}}QQP\\ (Acc)\end{tabular} & Avg. \\
\midrule
 Full-FT & 6.6B & $88.4$ & $88.7$ & $89.8$ & $67.9$ & $\underline{92.3}$ & $93.6$ & $86.3$ & $\textbf{91.7}$ & $87.3$ \\
 LoRA & 4.2M & $\underline{89.2}_{\pm 0.5}$ & $\underline{89.7}_{\pm 0.5}$ & $\underline{89.9}_{\pm 0.1}$ & $\textbf{70.6}_{\pm 0.7}$ & $\textbf{96.8}_{\pm 0.2}$ & $\underline{94.7}_{\pm 0.2}$ & $\textbf{90.9}_{\pm 0.2}$ & $\underline{91.6}_{\pm 0.1}$ & $\underline{89.2}$ \\
 LoRA-drop (ours) & 2.2M & $\textbf{91.0}_{\pm 0.5}$ & $\textbf{90.2}_{\pm 0.3}$ & $\textbf{90.1}_{\pm 0.1}$ & $\underline{69.0}_{\pm 1.2}$ & $\textbf{96.8}_{\pm 0.2}$ & $\textbf{94.8}_{\pm 0.2}$ & $\underline{90.6}_{\pm 0.1}$ & $\underline{91.6}_{\pm 0.3}$ & $\textbf{89.3}$ \\
\bottomrule
\end{tabular}}
\caption{The performance of the Llama2-7b on GLUE benchmark.}
\label{tab:llama}
\end{table*}

\begin{table*}[t!]
\centering
\scalebox{0.8}{
\begin{tabular}{l|c|cccc}
\toprule
Model & \#Tr. & \multicolumn{4}{c}{Dialogsum} \\
\multicolumn{1}{l|}{Llama2 7b} & \multicolumn{1}{l|}{Params} & \multicolumn{1}{l}{ROUGE-1} & \multicolumn{1}{l}{ROUGE-2} & \multicolumn{1}{l}{ROUGE-L} & \multicolumn{1}{l}{Avg.} \\ 
\midrule
Full-FT          & 6.6B        & $49.86$           & $\textbf{29.37}$   & $43.07$           & $40.77$          \\
LoRA             & 0.13B       & $\textbf{50.15}$  & $29.28$            & $\textbf{43.65}$  & $\textbf{41.03}$ \\
LoRA-drop (ours) & 0.09B       & $49.84$           & $28.99$            & $43.22$           & $40.68$          \\ 
\bottomrule
\end{tabular}}
\caption{Results of Llama2-7b with different training strategies on the summarization dataset Dialogsum. }
\label{tab:summarization}
\end{table*}

\begin{figure*}
    \centering
    \includegraphics[width=\linewidth]{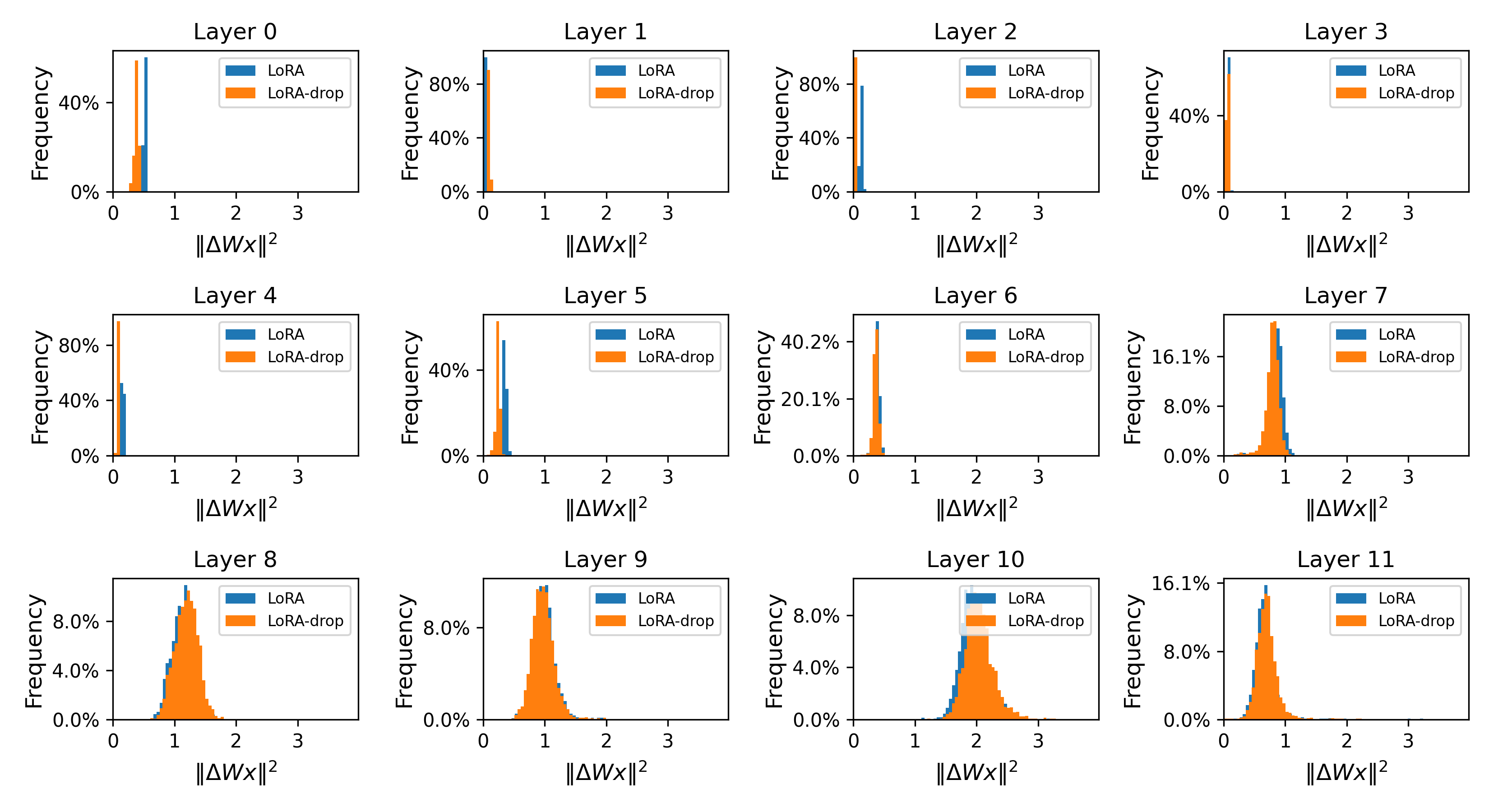}
    \caption{The query LoRA output $\Delta \bm{W}_i\bm{x}_i$ squared norm frequency distribution of LoRA and LoRA-drop.}
    \label{fig.rte_query_share}
\end{figure*}

\begin{figure*}
    \centering
    \includegraphics[width=\linewidth]{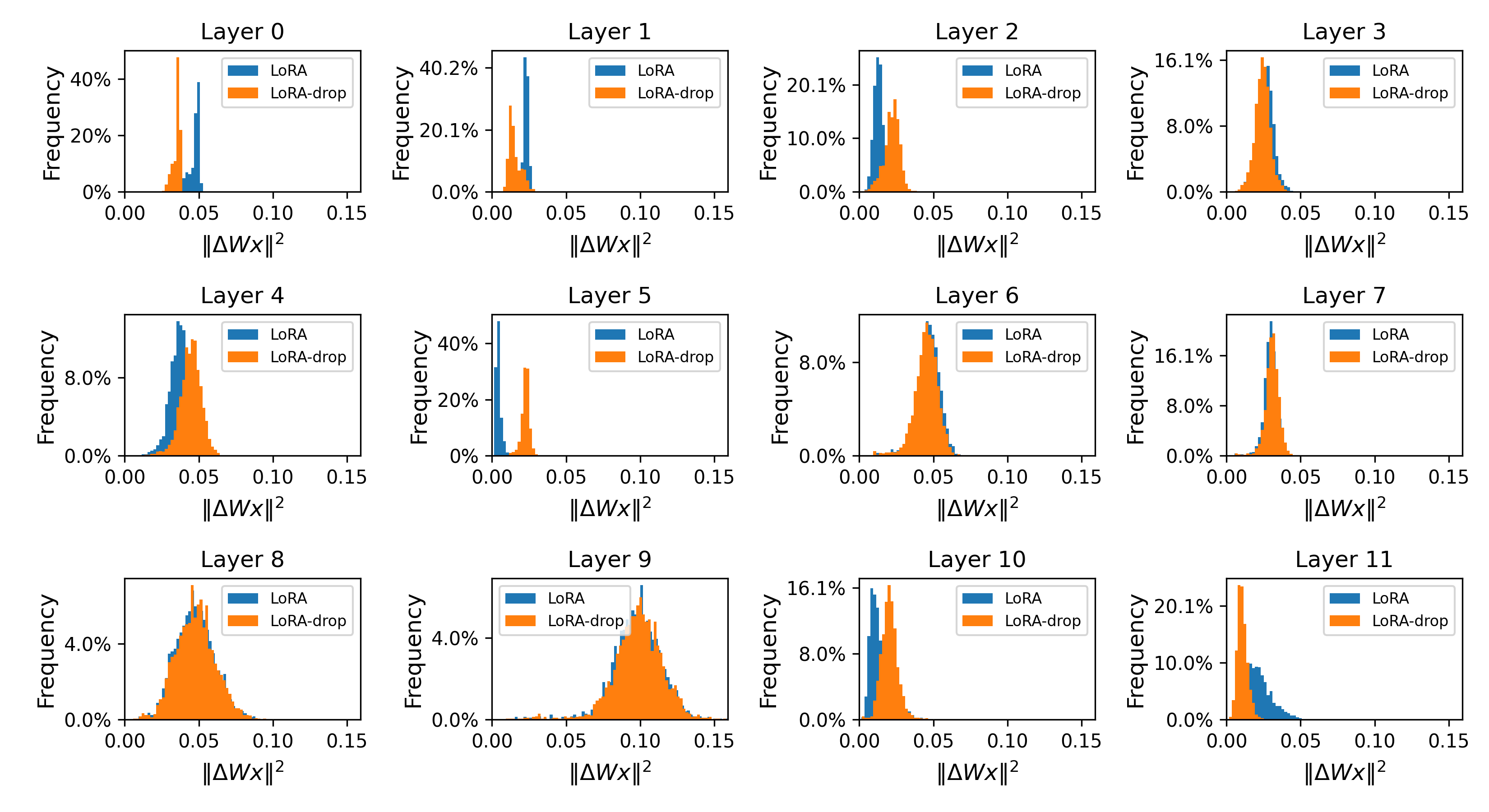}
    \caption{The value LoRA output $\Delta \bm{W}_i\bm{x}_i$ squared norm frequency distribution of LoRA and LoRA-drop.}
    \label{fig.rte_value_share}
\end{figure*}

\begin{figure*}[h]
    \centering
    \includegraphics[width=\linewidth]{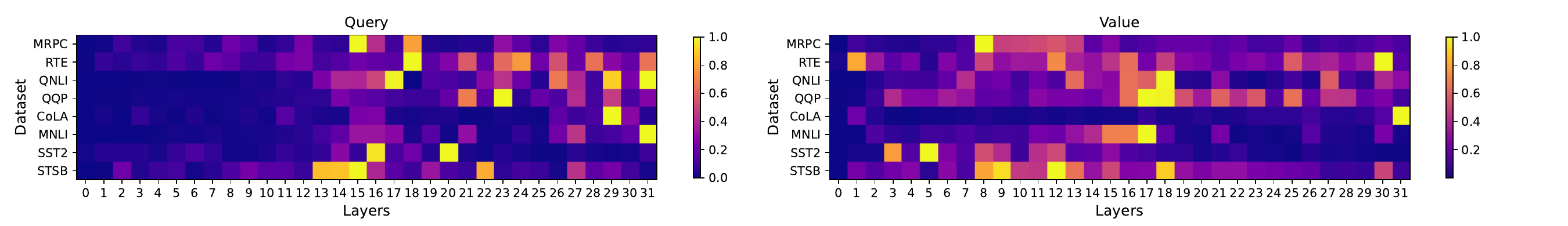}
	\caption{The relative magnitudes of LoRA outputs across different layers of Llama2-7b on various datasets. The left subplot shows the LoRA outputs corresponding to each layer's query matrix, and the right subplot shows the LoRA outputs corresponding to each layer's value matrix. For display, the value of the largest layer's LoRA output is normalized to 1 for each dataset.}
	\label{fig:llama different datasets}
\end{figure*}

\begin{figure}[h]
    \centering
    \includegraphics[width=\linewidth]{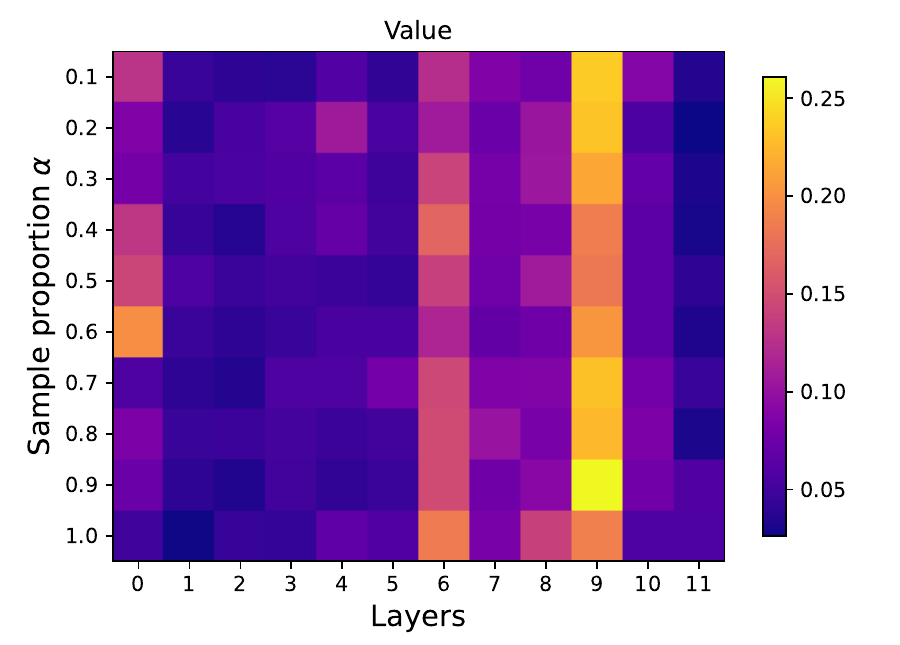}
	\caption{Importance distribution of LoRA for value in RTE under different sample proportions. Each point on the heatmap represents the importance $I_{i}$ of the query value in layer $i$ under $\alpha$ sample proportion. }
	\label{fig.sample}
\end{figure}

\end{document}